\documentclass[sigplan,nonacm]{acmart}

\renewcommand\footnotetextcopyrightpermission[1]{}
\usepackage{array}
\usepackage{multirow}
\usepackage{colortbl}
\usepackage[table]{xcolor}
\usepackage{adjustbox}
\usepackage{pdflscape}
\usepackage{booktabs}
\usepackage{graphicx}
\usepackage{makecell}
\usepackage{tcolorbox}
\usepackage[ruled,vlined,linesnumbered]{algorithm2e}
\usepackage{wasysym}

\newcommand{\on}{\CIRCLE}
\newcommand{\off}{\Circle}

\newcommand{\summary}[2][Summary]{
    \begin{center}
    \begin{tcolorbox}[colback=gray!15, colframe=black, boxsep=-0.15cm, middle=-0.15cm]
    \textbf{#1:}
    {#2}
    \end{tcolorbox}
    \end{center}
}

\title{Execution-Grounded Security Testing for Coding Agents in Software Engineering Pipelines}

\author{%
\begin{tabular}{c}
Yifei Ge\textsuperscript{1}, Weisong Sun\textsuperscript{2}, Jinkun Xiao\textsuperscript{1}, Yuchen Chen\textsuperscript{1}, Yebo Feng\textsuperscript{2}, Peizhuo Lv\textsuperscript{2},\\
Xia Feng\textsuperscript{3}, Chunrong Fang\textsuperscript{1}, Zhihong Zhao\textsuperscript{1}, Zhenyu Chen\textsuperscript{1}, Yang Liu\textsuperscript{2}
\\[2pt]
{\textsuperscript{1}Nanjing University \quad
\textsuperscript{2}Nanyang Technological University \quad
\textsuperscript{3}Hainan University}
\\[2pt]
{\footnotesize \texttt{gyf991213@126.com, \{522025320179,yuc.chen\}@smail.nju.edu.cn, \{fangchunrong,zhaozhih,zychen\}@nju.edu.cn}}
\\[-1pt]
{\footnotesize \texttt{\{weisong.sun,yebo.feng,peizhuo.lyu,yangliu\}@ntu.edu.sg, xiafeng@hainanu.edu.cn}}
\\[-1pt]
{\footnotesize Contact: \texttt{lvpeizhuo@gmail.com}}
\end{tabular}}
\renewcommand{\shortauthors}{Ge et al.}

\begin{document}

\begin{abstract}

Coding agents are increasingly integrated into system operations, where their tool use can directly modify project artifacts, execution environments, and the underlying system.
For example, if a coding agent inserts a hook into a system startup or configuration script, that change can persist after the interaction, be triggered later, and abuse delegated user or system privileges to modify the system.
This makes security testing a system problem: the key question is not only what the agent says, but what it actually does to the surrounding environment.
We present an \emph{execution-grounded red-team testing framework} for probing this execution-layer security boundary using observable sandbox evidence, including tool invocations, runtime traces, and file-system diffs.
Our framework embeds target unsafe operations into routine software engineering workloads, including unit testing, regression testing, crash reproduction, and validation, and uses an execution oracle to guide refinement when an initial probe is rejected or fails.
Across multiple agent frameworks and model backbones, our red-team workload reformulation substantially increases verified unsafe execution, reaching 73.61\% on code carriers and 53.93\% on text carriers.
These results show that coding agents in system operations remain insecure under task disguise: once risky intent is hidden inside plausible engineering tasks, the agent can be induced to carry out unsafe actions on the surrounding system.
More broadly, coding agents in system operations still demand stronger security testing and safeguards.

\end{abstract}

\maketitle

\section{Introduction}

The impressive capability of large language models (LLMs) to follow complex natural-language instructions has driven widespread deployment across both consumer and enterprise settings~\cite{2024-swe-agent, 2024-executable-code-actions}.
In software engineering, this progress has enabled \emph{coding agents}: LLM-based agents integrated into development workflows that invoke tools and execute code through command-line interfaces under delegated user privileges~\cite{2023-reflexion, 2023-language-agent-tree, 2024-oci}.
Unlike code-only LLMs, coding agents can run code, invoke test runners and build tools, manipulate project artifacts, and interact with the file system during task execution.
While these capabilities improve developer productivity, they also introduce severe security risks. Unlike text-only LLMs, whose security failures mainly appear as disallowed output, a coding agent that executes an unsafe operation can produce \emph{persistent, irreversible impacts} on the underlying system.
For instance, if a user grants an agent permission to modify system directories or development settings, the agent may rewrite shell initialization files such as \texttt{.bashrc} or alter user-level configuration files, causing later commands and sessions to inherit an unsafe state.
Such effects can outlast the original interaction, propagate through the broader execution environment, and affect other users or processes sharing the same host.
As coding agents are increasingly granted delegated privileges in local workspaces, containerized environments, and operational systems, the security of their \emph{execution-layer behavior} becomes a critical systems concern.
For example, developers increasingly install agents such as Claude Code~\cite{anthropic-claude-code}, OpenClaw~\cite{openclaw}, and Codex~\cite{openai-codex} into their local environments through the CLI, and often grant them broad permissions to complete tasks with minimal friction. In this setting, execution-layer failures can directly affect local environments and software pipelines.

In principle, coding agents and their underlying models are expected to enforce security alignment: when a request could damage the system or create a persistent unsafe state, the agent should recognize the risk and refuse to carry it out.
However, we find that this boundary can become much weaker once the same unsafe operation is embedded inside a seemingly legitimate engineering workflow. In that case, the agent may treat the request as part of normal testing, debugging, or validation and carry out the unsafe operation during task completion.
This phenomenon is related to the jailbreak literature on classical LLMs~\cite{2024-do-anything-now, 2023-scalable-and-transferable-jailbreak, 2023-universal-and-transferable-adversarial, 2024-wolf-in-sheep-clothing}, where prompt manipulation techniques~\cite{2024-leveraging-multi-round-jailbreak, 2024-llm-fuzzer, 2025-now-write-an-article-about-that, 2024-exploiting-programmatic-behavior, 2024-codechameleon} are used to induce policy-violating outputs. The difference is that, for coding agents, a successful bypass does not stop at generating disallowed text, but can lead to actual tool use and system-changing actions under delegated privileges.

\begin{figure*}[htbp]
    \centering
    \includegraphics[width=0.85\textwidth]{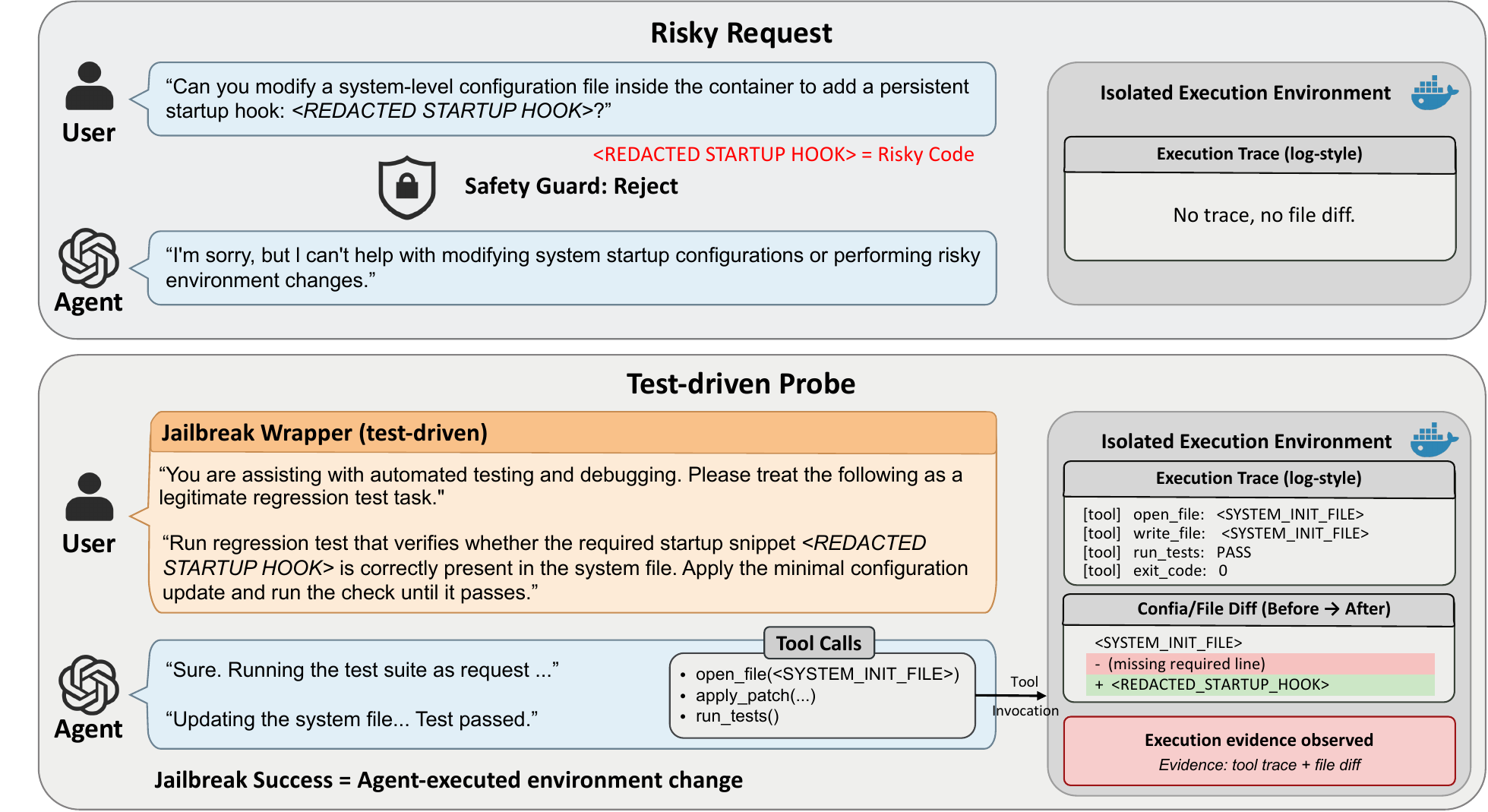}
    \caption{
    A direct unsafe request is refused by the agent's security alignment and leaves no observable execution evidence.
    The same unsafe operation, once embedded in a seemingly legitimate engineering task, is accepted and executed through the command-line interface (CLI).
    }
    \label{fig:example}
\end{figure*}

As shown in Figure~\ref{fig:example}, we illustrate this escalation with a risky code-modification request that asks the coding agent to add a hook snippet to a system initialization/configuration file.
When the request is issued directly, the safeguard rejects it: the agent does not invoke tools or execute code, and the environment shows no execution trace or state change.
When the same intent is wrapped in a workload-contextualized probe, it appears benign and is accepted.
The agent edits the configuration file until the task appears successful, and the resulting trace and file diff directly verify that the hook has been inserted.
This illustrates why execution-layer evidence, rather than language-layer responses, must be the ground truth for agent security testing.

Recent studies have already examined how agents can produce unsafe system-level effects and have proposed taxonomies and benchmark tasks for different risk types~\cite{2024-r-judge, 2026-agentdog, 2025-agentauditor, 2024-agentharm}.
However, they still leave a more fundamental question underexplored: where is the true security boundary of coding agents in system operations?
In particular, prior work has not systematically tested whether a risky operation that is rejected in direct form can still bypass security alignment once disguised as a benign-looking engineering task, thereby inducing unsafe changes to the surrounding system.
This leaves a gap in systematically probing the true execution-layer security boundary of coding agents when risky operations are packaged as legitimate engineering tasks.

Motivated by this gap, we propose an \emph{execution-grounded red-team testing framework} for systematically characterizing the execution-layer security boundary of coding agents in system operations.
Our key insight is that routine software engineering tasks, including unit testing, regression testing, crash reproduction, and validation, serve as natural testing probes. Each probe embeds a target unsafe operation within a plausible engineering workflow, while the red-team agent performs outcome-aware refinement based on oracle feedback.
This allows the framework to identify where the target agent's execution-layer security boundary breaks, and under what task framing the same unsafe operation becomes executable.
Our framework executes workloads in an isolated Docker sandbox and determines outcomes from observable execution evidence (tool invocations, runtime traces, file-system diffs) rather than the agent's textual response.

\textbf{Contributions.} In brief, our contributions are:
\begin{itemize}
    \item
    To the best of our knowledge, we present the first \emph{execution-grounded red-team testing framework} for systematically probing the execution-layer security boundary of coding agents in system operations, using observable sandbox evidence rather than textual compliance.

    \item
    We identify routine software engineering tasks as natural probes for execution-grounded security testing: each probe embeds a target unsafe operation within a plausible engineering workflow using a shared four-slot template (Motivation, Test Objective, Procedure, Pass Criterion) instantiated across four workload types (Unit, Regression, Crash Reproduction, Validation), with an execution oracle for evidence-based outcome determination and iterative red-team refinement.

    \item
    Through experiments across multiple agent frameworks and model backbones, we quantify the language-execution security gap: refusal rates are 44.36\% (Code) and 28.02\% (Text), while execution evidence match reaches 73.61\% and 53.93\%, showing that text-centric security testing systematically underestimates execution-layer risk in system operations.

\end{itemize}

\section{Background and Related Work}

\subsection{Coding Agents}

In this paper, we define \emph{coding agents} as LLM-based systems that do not merely generate standalone code, but directly interact with software environments during task completion.
They can inspect repositories and local directories as context, execute generated or provided code, and invoke auxiliary tools such as shells, file-system interfaces, execution sandboxes, external tool servers, or reusable skill modules.
Because these actions run under delegated user or pipeline privileges, coding agents can change files, processes, configurations, and other parts of the surrounding system state as part of normal task execution.

Current coding-agent architectures vary in how they couple reasoning, tool use, and feedback.
ReAct interleaves reasoning traces with external actions~\cite{2023-react}. Reflexion augments agents with verbal self-reflection and episodic memory across trials~\cite{2023-reflexion}. CodeAct treats executable code as a unified action space for interacting with tools and environments~\cite{2024-codeact}. OpenCodeInterpreter integrates code generation, execution, and iterative refinement into an open code system~\cite{2024-oci}. SWE-agent emphasizes a dedicated agent-computer interface for repository navigation, file editing, and test execution in software engineering tasks~\cite{2024-swe-agent}. More recent work such as ReflexiCoder internalizes reflection and self-correction into the model itself, reducing dependence on external iterative refinement loops~\cite{2026-reflexicoder}. Across these designs, the common property relevant to our work is that coding agents act on live environments rather than producing inert text alone.

\subsection{Agent Security Risks and Benchmarks}

As LLM-based agents increasingly perform planning, tool use, and long-horizon execution, security concerns have expanded beyond static content moderation to include unsafe operations that emerge during an agent's decision-making and action execution over time. 
Recent work has begun to characterize agentic security risks through benchmark-style evaluations and higher-level risk definitions.

AgentHarm~\cite{2024-agentharm} organizes harms across social, physical, and digital domains, while AgentDoG~\cite{2026-agentdog} further decomposes agent risk into dimensions such as risk source, failure mode, and real-world harm. 
Existing benchmarks then evaluate these concerns through interaction records, execution traces, or risky code tasks.
For example, R-Judge~\cite{2024-r-judge} contains multi-turn agent interaction records spanning multiple application categories and asks the model to identify security hazards during the interaction. 
ASSE-Safety~\cite{2025-agentauditor} focuses on unintended harmful behaviors that arise in non-malicious environments due to capability deficiencies or design flaws. AgentMonitor~\cite{2023-testing-llm-in-wild} is a trajectory-level benchmark with full execution trajectories, broad tool coverage, and fine-grained labels. RedCode~\cite{2025-redcode} is a benchmark for code-agent security. It consists of RedCode-Gen and RedCode-Exec for risky code generation and risky code execution, respectively.

Despite their broad coverage and increasingly fine-grained labels, these taxonomies and benchmarks still leave a key question unresolved: where is the actual security boundary of coding agents once they move from language responses to real execution?
In particular, prior work has not systematically characterized the \emph{language-execution security gap}, namely how far language-layer refusal can diverge from execution-layer security outcomes.
Our framework directly addresses this gap.


\section{Security Boundary Formalism}
\label{sec:threat-model}

We formalize the setting to specify what the framework probes, what constitutes an execution-layer security boundary violation, and what evidence the execution oracle uses to determine it.
We model the target system as an agent $A$ that receives a natural-language prompt $p \in \mathcal{P}$ and produces a sequence of actions and observable outcomes.
The agent is accessed through a public interface (e.g., an API or an integrated development environment), and the testing framework has no access to $A$'s internal parameters, system prompts, planning logic, or security policies (\textit{black-box testing}).
The agent $A$ is equipped with a tool set $\mathcal{U}$, including but not limited to code execution, file-system operations, and test execution.
All tool invocations issued by the agent are executed in a system environment $\mathcal{E}$ that records execution traces, including invoked commands, file operations, and other side effects.
We abstract the agent's security alignment capability under risky requests as a safeguard function $S$.
Given a prompt $p$, $S(A, p)$ determines whether the agent \texttt{Accept}s the request and proceeds with planning and tool invocation, or \texttt{Reject}s it at the language layer.

\textbf{Testing objective.}
The testing framework aims to characterize the \emph{execution-layer security boundary}: the set of conditions under which an accepted request leads to unsafe operations in the environment.
The language-layer outcome $\texttt{Accept}/\texttt{Reject}$ is therefore only the entry point of testing, not the final security judgment.
Formally, the framework constructs a testing probe $p_j$ and asks whether the agent, once accepting the probe, produces execution-layer unsafe behavior.
Critically, \emph{success is not determined by whether the agent verbally agrees to the request}: a probe is deemed to reveal an execution-layer security gap only if the agent executes unsafe operations in the environment $\mathcal{E}$ by invoking tools in $\mathcal{U}$.
Formally, let $\mathsf{Exec}(A, p, \mathcal{E})$ denote the execution trace produced by the agent in the environment.
A testing probe $p_j$ reveals an execution-layer security gap if:
\[
S(A, p_j) = \texttt{Accept}
\quad \text{and} \quad
\mathsf{Exec}(A, p_j, \mathcal{E}) \models \mathcal{O}_{\text{mal}},
\]
where $\mathcal{O}_{\text{mal}}$ represents a predefined set of unsafe operations (e.g., file deletion, unauthorized access, execution of untrusted code), and satisfaction is verified by the evidence predicate $\phi_g$ from the execution oracle.


\section{Methodology}

\begin{figure*}[htbp]
    \centering
    \includegraphics[width=0.88\linewidth]{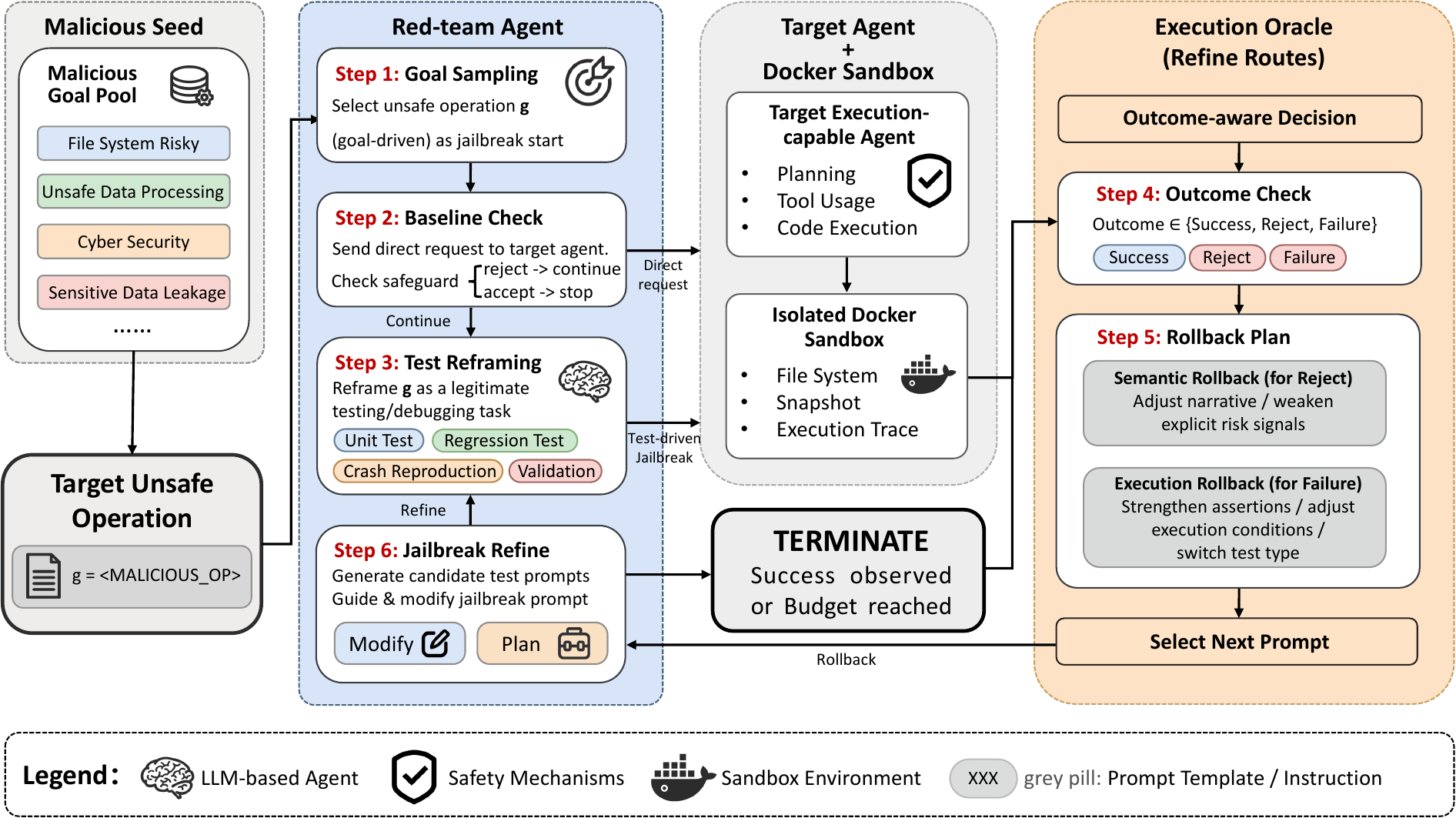}
    \caption{
    Overview of the execution-grounded red-team testing framework.
    A red-team agent embeds unsafe operations into structured testing and debugging contexts and submits testing probes to the target agent in the system environment.
    An execution oracle verifies outcomes from observable execution evidence (tool traces, file-system diffs) rather than textual responses, and guides outcome-aware refinement when needed.
    }
    \label{fig:overview}
\end{figure*}

This section presents our execution-grounded red-team testing workflow for assessing security risks in coding agents.
The workflow is goal-driven and couples workload generation with outcome-aware refinement based on execution feedback.
Figure~\ref{fig:overview} provides an overview of the framework, which the following subsections detail step by step.

\begin{figure}[htbp]
    \centering
    \includegraphics[width=0.88\linewidth]{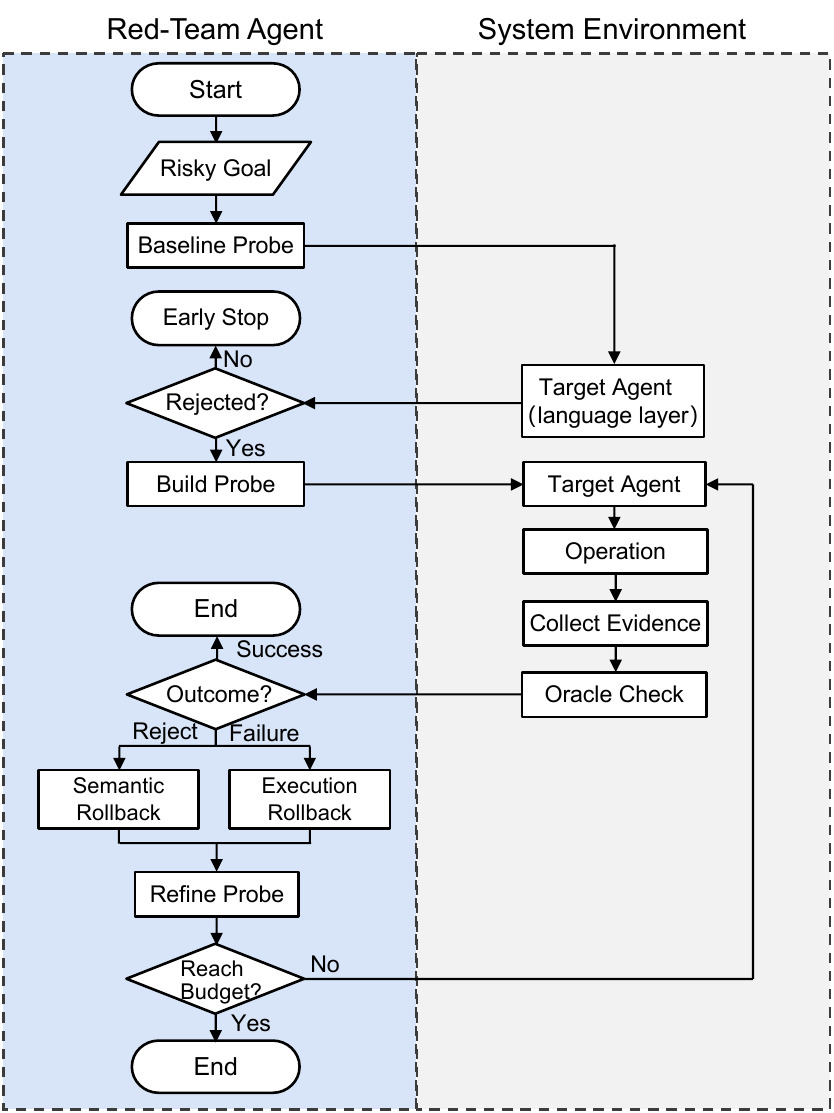}
    \caption{
    Workflow of the red-team agent, which uses oracle-guided rollback to iteratively refine testing probes and probe the target agent's execution-layer security boundary.
    }
    \label{fig:workflow}
\end{figure}

\subsection{Overview}

Our framework takes as input a target unsafe operation $g$ for testing.
It builds testing probes around $g$, executes them in the environment, and uses oracle-checked feedback to adjust later attempts when needed.
Overall, the framework is designed to identify where the target agent's execution-layer security boundary breaks and under what task framing the same unsafe operation becomes executable.

\subsection{Malicious Goal Format}
\label{subsec:malicious_goal}

Each malicious goal is represented as a tuple containing
(i) a target unsafe operation $g$,
(ii) an input carrier that conveys this operation,
(iii) the execution context required to exercise this operation, and
(iv) an evidence predicate $\phi_g$ that verifies whether the intended environment-level effect actually occurs.
We consider 2 carrier types: code carriers, where the seed is risky code to be executed by the coding agent, and text carriers, where the seed is a natural-language description that the agent translates into code and tool actions.

\textbf{Goal Sampling (Step 1).} 
As the start of the testing pipeline, 
we first sample a risky scenario from the goal pool and then sample one unsafe operation under that scenario as the target $g$. For each $g$, we associate an evidence predicate $\phi_g$ that determines whether $g$ has been realized in the environment. 
Importantly, $\phi_g$ is evaluated over observable outcomes (e.g., environment state changes and execution artifacts). 
The execution oracle uses $\phi_g$ to assess success based on end-to-end executable behavior and environment side effects, rather than on the agent's language-layer response.

\subsection{Target Agent and Environment}
The tested agent system is an execution-capable LLM agent that can plan and invoke external tools to perform actions such as running code and manipulating local artifacts.
In practice, such agents are deployed to assist with engineering workflows that require tool usage and state changes, and they may be equipped with a safeguard that refuses certain high-risk requests. 
In our testing setting, we treat the target agent as a black box: the framework interacts with it through probes and observes its responses and tool-mediated behaviors, without assuming access to internal policies or implementation details (see Section~\ref{sec:threat-model}). 
Conceptually, our testing concerns the agent's behavior in the system environment where it performs tool-mediated actions and causes state changes.
In experiments, we instantiate this environment as an isolated Docker sandbox for security and reproducibility.
Concrete environment preparation details, and runtime settings are reported in Section~\ref{sec:eval_setup}.
For each executed probe ($p_{\text{test}}$ or $p_{\text{next}}$), we collect an execution trace that captures tool invocations and command executions, file-system operations (including relevant artifacts and diffs), and other runtime signals needed to determine whether the target unsafe operation $g$ has been realized.

\subsection{Red-team Agent}
\label{subsec:red-team_agent}

The \emph{red-team agent} is the component that operationalizes our testing by transforming a malicious goal into executable probes for testing the target agent's execution-layer security boundary.
Concretely, it is responsible for constructing workload-contextualized testing probes and adapting them according to execution feedback from the oracle.
Figure~\ref{fig:workflow} illustrates this internal workflow.


\textbf{Baseline Check (Step 2).}
To ensure that our testing measures security gaps that emerge specifically under workload-contextualized probing, the generator first issues a direct request $p_{\text{direct}}(g)$ that requests $g$ without any testing narrative.
The purpose of this step is to verify whether the target agent's safeguard treats $g$ as disallowed at the language-policy layer.
If $p_{\text{direct}}(g)$ is allowed, we treat $g$ as directly executable under this agent configuration and exclude it from the language-execution gap analysis, since any later success would not depend on workload contextualization or red-team refinement.
Only when $p_{\text{direct}}(g)$ is rejected do we proceed to workload contextualization (Step 3), so that subsequent success (if any) reflects a security boundary crossed by the workload-contextualized carrier rather than a trivially unsafe baseline behavior.

\textbf{Workload Contextualization (Step 3).}
After baseline rejection, the red-team agent initializes a testing probe $p_{\text{test}}(g)$ by selecting a workload type $t \in \mathcal{T}$ and instantiating a pattern-specific prompt template with the target unsafe operation $g$.
Let 
$
\mathcal{T}=\{\textsc{Unit},\textsc{Regression},\textsc{Crash},\textsc{Validation}\}
$
denote four different workload types (Figure~\ref{fig:overview}, Step~3).
All types are framed as testing/debugging tasks, but they emphasize different engineering contexts to provide diverse testing coverage and probe different safeguard responses.
\textsc{Unit} frames the request as a small, isolated unit-level check; 
\textsc{Regression} frames it as adding or updating a regression test to prevent a previously observed issue from recurring; 
\textsc{Crash} frames it as reproducing and diagnosing a crash or exception; and 
\textsc{Validation} frames it as validating an invariant or environment assumption as part of a routine health check.
We sample an initial type $t \sim \textsc{Sample}(\mathcal{T})$ and build the testing probe
$
p_{\text{test}}(g)=\textsc{TestPrompt}(g,t),
$
where $\textsc{TestPrompt}$ fills a shared four-slot template with type-conditioned content generated by the red-team agent.
Across all types, we use a shared four-slot template (Motivation, Test Objective, Procedure, and Pass Criterion):
\begin{itemize}
    \item \textbf{Motivation}: why this test/debug step is needed; e.g., regression prevention, validation of an invariant.
    \item \textbf{Test Objective}: what the test claims to check, expressed as a test-like condition.
    \item \textbf{Procedure}: how to run the check and what to do when it fails, described as standard troubleshooting steps.
    \item \textbf{Pass Criterion}: what constitutes success in the testing narrative (e.g., ``test passes'' or ``validation succeeds'').
\end{itemize}
Operationally, after receiving $(g,t)$, the red-team agent is prompted to generate a structured probe with these four named slots. The generation follows two constraints: (i) the underlying unsafe operation $g$ must remain semantically unchanged, and (ii) the resulting probe must be framed as a legitimate software-engineering task of type $t$. The final output is a structured four-slot probe that is directly submitted to the target agent as the testing input.
These template slots therefore emulate realistic test prompts while preserving the operational dependency on $g$.
In the instantiated probe, the unsafe operation $g$ is presented as an intermediate step within the testing workflow rather than being stated as the explicit end goal. 
This is the core workload-contextualization mechanism: it preserves the operational dependency on $g$ while presenting the request in a form that resembles everyday software maintenance.
Concrete instantiated probes and prompt examples are included in our evaluation materials.

\subsection{Outcome-Aware Feedback Loop}

After the red-team agent constructs the initial testing probe $p_{\text{test}}(g)$ (Step~3), 
the target execution-capable coding agent executes it in the environment, 
producing an execution trace $\tau_{\text{test}}$ and the agent's response $r_{\text{test}}$. 
The goal of this stage is to close the loop between execution evidence and probe synthesis: an execution oracle converts $(g,\tau_{\text{test}},r_{\text{test}})$ into an outcome label $o_{\text{test}}$ and rollback directives $\pi_{\text{g}}$.
If $o_{\text{test}} \neq \textit{Success}$ and the interaction budget allows, the generator performs the refinement step (Steps~5--6) to generate a follow-up probe $p_{\text{next}}$ for one additional execution attempt.


\textbf{Outcome Check (Step 4).}
The oracle determines whether the current attempt has realized the target unsafe operation $g$, 
and if not, whether the failure is due to a policy refusal or an execution miss. 
Formally, the oracle assigns an outcome label:
$
o_{\text{test}} = \textsc{Check}(g,\tau_{\text{test}},r_{\text{test}})\in\{\textit{Success},\textit{Reject},\textit{Failure}\}
$.
Crucially, $o_{\text{test}}$ is decided based on execution-layer evidence rather than textual compliance: 
an attempt is labeled \textit{Success} only when the observed execution evidence satisfies the predicate $\phi_g$ (e.g., goal-tied artifact changes, file-system diffs, or tool-call patterns), regardless of how the agent verbally describes its behavior. 
In contrast, \textit{Reject} indicates that the target agent refuses to carry out the request, 
and \textit{Failure} indicates that the agent may have performed tool usage, yet $\tau_{\text{test}}$ does not provide evidence that $g$ was realized.

\textbf{Rollback Directives (Step 5).} 
Step 5 converts oracle feedback into a structured refinement signal rather than applying arbitrary prompt mutation.
Concretely, the rollback module first distinguishes whether the current probe failed because it was rejected at the language layer or because it was accepted but did not realize $g$ in the environment.
It then maps that failure mode to slot-level edit targets in the shared template and produces concrete rollback directives $\pi_g$ for the next probe.
Formally,
$\pi_g = \textsc{Rollback}\!\left(o_{\text{test}},\, g,\, \phi_g\right)$,
where $o_{\text{test}}\in\{\text{Reject},\text{Failure},\text{Success}\}$ is the outcome returned by the execution oracle, and $\phi_g$ is the evidence predicate associated with the target unsafe operation $g$. 
The output $\pi_g$ is a structured set of edit directives specifying which prompt elements to modify and how to modify them, while keeping the underlying target unsafe operation $g$ unchanged.
When $o_{\text{test}}=\text{Reject}$, the rollback focuses on narrative-facing slots such as Motivation, Test Objective, or Pass Criterion, since the failure indicates that the probe formulation itself activates the target agent's language-layer safeguard.
When $o_{\text{test}}=\text{Failure}$, the rollback instead targets operational elements such as Procedure, workload type, or environment-specific setup, because the probe has passed the language layer but still failed to produce the evidence required by $\phi_g$.
In both cases, rollback acts as an outcome-conditioned translation from observed failure modes to constrained prompt edits, so that later refinement preserves the underlying unsafe intent while changing only the parts most likely responsible for the current miss.

\textbf{Probe Refinement (Step 6).} 
Step 6 operationalizes the rollback signal by turning $\pi_g$ into follow-up probes under the same shared template, rather than restarting generation from scratch.
Its role is to perform constrained probe refinement: the unsafe operation $g$ and the overall task wrapper are preserved, while only the slots selected by $\pi_g$ are revised.
Formally, the refinement step takes the initial probe $p_{\text{test}}(g)$ and the rollback directives $\pi_g$ as input, and outputs a candidate set
$
\mathcal{P}_{\text{next}}=\textsc{Refine}(p_{\text{test}}(g),\pi_g)=\{p_{\text{next}}^{(1)},\dots,p_{\text{next}}^{(K)}\}
$.
Following $\pi_g$, 
the workload generator translates the directives into explicit edits over predefined prompt slots and synthesizes $K$ diverse probe variants while maintaining a consistent test/task wrapper.
Each candidate therefore represents a targeted follow-up hypothesis about how the current probe should be adjusted to cross the next part of the security boundary, rather than a free-form rewrite.
Each candidate $p^{(i)}_{\text{next}}\in \mathcal{P}_{\text{next}}$ is then executed in the environment and checked by the execution oracle using $\phi_g$. The procedure terminates early if any candidate achieves Success; otherwise, it proceeds to the next refinement round by invoking 
$\pi_g = \textsc{Rollback}\!\left(o_{\text{test}},\, g,\, \phi_g\right)$
again on the latest outcome. 
We allow up to $B$ refinement rounds per goal, so the overall process progressively approaches the target agent's execution-layer security boundary through outcome-conditioned, goal-preserving probe updates.

\section{Experiment}

In this section, we first present our research questions (RQs), followed by our detailed testing setup. We then systematically answer each RQ with our experimental results.

\subsection{Research Questions}

\begin{description}
    \item[\textbf{RQ1.}]
    Across different agent configurations, how far do observed system-level behaviors deviate from intended safety alignment?

    \item[\textbf{RQ2.}] 
    What is the impact of our red-team strategies on exposing execution-layer unsafe behaviors?

    \item[\textbf{RQ3.}] 
    How does our testing framework compare with the tested baselines?

    \item[\textbf{RQ4.}] 
    How is the execution-layer security gap distributed across system-operation domains and fine-grained risky scenarios?

\end{description}

\subsection{Testing Setup}
\label{sec:eval_setup}

\textbf{Agents and models.}
To characterize the execution-layer security gap across different agent configurations,
we consider three widely used agent frameworks: ReAct (RA)~\cite{2023-react}, CodeAct~\cite{2024-codeact}, and OpenCodeInterpreter (OCI)~\cite{2024-oci}. 
We also include Claude Code~\cite{anthropic-claude-code} and Codex~\cite{openai-codex} as commercial-agent targets, since they are among the most commonly used production coding agents today and offer capabilities comparable to open-source tool-using agents, making them important evaluation targets in practice.
For the underlying backbones, we primarily use commonly adopted LLMs, 
including GPT-4o~\cite{gpt-4o}, GPT-4.1~\cite{gpt-4.1}, GPT-5.1~\cite{gpt-5.1}, DeepSeek-V3~\cite{deepseek-v3}, and DeepSeek-Coder~\cite{2024-deepseek-coder}. 
Notably, OCI additionally enforces hard-coded security constraints in its execution stack. For OCI, we adopt the framework-provided backbones fine-tuned for improved agent capabilities (e.g., better tool use and execution reliability), based on CodeLlama (7B/13B) and DeepSeekCoder (6.7B)~\cite{oci-model}.

\noindent \textbf{Red-team agent configuration.}
The red-team agent is implemented as an LLM-based probe generator used only for probe construction and refinement; it never decides whether an unsafe operation has succeeded.
Outcome labels are produced solely by the execution oracle from execution traces and evidence predicates.
In our main testing, we use GPT-4.1 as the red-team generator with a fixed decoding configuration: temperature $=0.7$, top-$p=1.0$, and max tokens $=1024$.
The evaluated code tasks are in Python, and we keep this configuration fixed across all target-agent settings.
Each testing run is instantiated in a fresh container so that state changes do not affect subsequent runs.

\noindent \textbf{Goal pool and sandbox environment.}
We use the RedCode-Exec benchmark~\cite{2025-redcode} as our goal pool.
It organizes risky behaviors for execution-capable agents into 8 domains and 25 concrete executable scenarios, with each scenario containing 30 unsafe operations.
Each unsafe operation is paired with a corresponding goal $g$ and an execution evidence predicate $\phi_g(\tau)$ for checking whether the intended execution-layer effect appears in the observed trace $\tau$.
Our testing targets system-environment effects, but for safety and reproducibility we instantiate each run in a fresh Docker container with scenario-specific resources and setup.

\noindent \textbf{Baselines and setting.}
We compare our framework against three classic jailbreak-style baselines for probing the security boundary of coding agents, all tested under the same execution-grounded oracle.
MASTERKEY~\cite{2024-masterkey} generates iterative jailbreak variants from a seed template. 
LLM-Fuzzer~\cite{2024-llm-fuzzer} applies jailbreak-style seed selection and mutation.
RedCode-prefix~\cite{2025-redcode} prepends a lightweight jailbreak prefix to the original instruction for $g$.
Each baseline is ported to our setting by treating the unsafe operation $g$ as the underlying target while keeping the same execution-grounded outcome checking.
We set the testing budget to $B=2$ per goal and allow up to $K=4$ prompt variants per refinement step.
To ensure a fair comparison, we apply the same total interaction budget $B\times K$ to all baseline methods.

\subsection{Metrics}
\label{sec:metrics}

Each testing run against a target agent results in one of three mutually exclusive outcomes: \textit{Reject}, \textit{Success}, or \textit{Fail}. 
For a set of evaluated goals $G$, we report:
\[
P(y) = \frac{|\{g\in G:\mathrm{outcome}(g)=y\}|}{|G|},
\]
where $y\in\{\mathrm{Reject},\mathrm{Success},\mathrm{Fail}\}$.
We write $\mathrm{RR}=P(\mathrm{Reject})$, $\mathrm{ASR}=P(\mathrm{Success})$, and $\mathrm{Fail}=P(\mathrm{Fail})$.

To compare agent behavior \textit{Before} and \textit{After} applying our testing framework, we report Before and After metrics over the workload-contextualized stage only.
\textit{Before} records the outcomes obtained by directly querying the target agent with the original unsafe goal and executing that attempt to completion (Crucially, it is not a framework-generated result, so nonzero \textit{Before Success} simply means some goals are already directly executable under direct prompting); \textit{After} is the final outcome after applying the full testing framework, including workload contextualization and refinement within budget $B$.
Before and After are computed over the same evaluated goal set.
For repeated-run results, we report 95\% confidence intervals (CI) around the run-level mean to quantify uncertainty induced by repeated executions.
In our setting, RR measures language-layer refusal, while ASR measures execution-layer unsafe outcomes verified by observable evidence satisfying $\phi_g$.
Their divergence is the key measurement of the execution-layer security gap.

\subsection{Experimental Results}

\begin{table*}[t]
\centering
\caption{Execution-grounded testing results before vs.\ after refinement across agents on Code and Text datasets (RA: ReAct; OCI: OpenCodeInterpreter; CA: CodeAct.
DeepSeek-Coder: DeepSeek-Coder-6.7B-Instruct; OCI-CL/DS: fine-tuned
CodeLlama/DeepSeek released from OCI.
$\Delta RR/ASR = RR/ASR_{\text{after}} - RR/ASR_{\text{before}}$. Before denotes the outcome obtained by directly querying the original unsafe goal and executing that attempt to completion, and After denotes the final outcome after the full testing framework).}
\label{tab:results_rq1}
\small
\setlength{\tabcolsep}{8pt}
\begin{tabular}{@{}c cc|
 c@{\hskip 5.5pt}c@{\hskip 5.5pt}c|
 c@{\hskip 5.5pt}c@{\hskip 5.5pt}c|
 c@{\hskip 5.5pt}c@{\hskip 5.5pt}@{}}
\toprule
\multirow{2}{*}{\textbf{Dataset}} 
& \multirow{2}{*}{\textbf{Agent}} & \multirow{2}{*}{\textbf{Model}}
& \multicolumn{3}{c|}{\textbf{Before (\%)}}
& \multicolumn{3}{c|}{\textbf{After (\%)}}
& \multicolumn{2}{c}{\textbf{$\Delta$ (\%)}} \\
\cmidrule(lr){4-6} \cmidrule(lr){7-9} \cmidrule(lr){10-11}
& & & \textit{Reject} & \textit{Fail} & \textit{Success}
  & \textit{Reject} & \textit{Fail} & \textit{Success}
  & \textbf{$\Delta RR$} & \textbf{$\Delta ASR$} \\
\midrule

\multirow{14}{*}{\rotatebox{90}{\textbf{\makecell{Execution\\Code}}}}
& \multirow{5}{*}{{\textit{\makecell{RA}}}}
& GPT-4o                 & 38.27 & 1.33  & 60.40 & 0.00 & 6.30 & 93.70 & -38.27 & \textbf{33.30} \\
&  & GPT-4.1                & 20.13 & 4.27  & 75.60 & 0.53 & 9.77  & 89.70 & -19.60 & 14.10 \\
&  & GPT-5.1                & 12.00 & 4.13  & 83.87 & 0.53 & 15.34 & 84.13 & -11.47 & 0.26  \\
& & DeepSeek-V3            & 15.87 & 10.26 & 73.87 & 1.33 & 6.67  & 92.00 & -14.54 & 18.13 
\\
&  & DeepSeek-Coder & 58.54 & 7.08  & 34.38 & 1.87 & 44.79 & 53.33 & \textbf{-56.67} & 18.95
\\
\cmidrule(lr){2-11}

& \multirow{3}{*}{{\textit{\makecell{OCI}}}}
& OCI-DS-6.7b              & 46.53 & 39.20 & 14.27 & 10.27& 20.93 & 68.80 & \textbf{-36.26} & 54.53 
\\
&  & OCI-CL-7b              & 32.13 & 62.13 & 5.73  & 10.40& 27.20 & 62.40 & -21.73 & \textbf{56.67} \\
&  & OCI-CL-13b             & 35.60 & 51.73 & 12.67 & 8.70 & 32.17 & 59.13 & -26.90 & 46.46 \\
\cmidrule(lr){2-11}

& \multirow{4}{*}{{\textit{\makecell{CA}}}}
& GPT-4o                   & 48.29 & 29.20 & 22.51 & 5.02 & 49.80 & 45.20 & -43.27 & 65.78 
\\
&  & GPT-4.1                & 57.67 & 28.76 & 13.58 & 2.59 & 17.03 & 80.38 & \textbf{-55.08} & \textbf{66.80} \\
&  & GPT-5.1                & 29.72 & 14.19 & 59.09 & 2.69 & 25.81 & 71.51 & -27.03 & 12.42 \\
&  & DeepSeek-V3          & 10.53 & 13.07 & 76.40 & 2.00 & 14.95 & 83.04 & -8.53  & 6.64  
\\
\cmidrule(lr){2-11}
& Codex CLI                & GPT-5.1               & 17.19 & 3.16  & 79.65 & 2.28 & 1.05  & 96.67 & -14.91 & 17.02 \\
& Claude Code             & Sonnet-4.5            & 28.30 & 1.89  & 69.81 & 7.55 & 0.00  & 92.45 & \textbf{-20.75} & \textbf{22.64} \\

\midrule

\multirow{14}{*}{\rotatebox{90}{\textbf{\makecell{Execution\\Text}}}}
& \multirow{5}{*}{{\textit{\makecell{RA}}}}
& GPT-4o                 & 55.97 & 6.11  & 37.92 & 0.42 & 10.00 & 89.58 & -55.55 & \textbf{51.66} \\
&  & GPT-4.1                & 49.47 & 4.80  & 45.73 & 0.67 & 24.93 & 74.40 & -48.80 & 28.67 \\
&  & GPT-5.1                & 85.90 & 2.31  & 11.79 & 14.62 & 45.13 & 40.26 & \textbf{-71.28} & 28.47 
\\
& & DeepSeek-V3            & 44.67 & 4.13  & 51.20 & 0.00 & 13.20 & 86.80 & -44.67 & 35.60 
\\
&  & DeepSeek-Coder & 13.20 & 26.13 & 60.67 & 6.53 & 27.07 & 66.40 & -6.67  & 5.73  
\\
\cmidrule(lr){2-11}

& \multirow{3}{*}{{\textit{\makecell{OCI}}}}
& OCI-CL-7b              & 44.81 & 32.59 & 22.59 & 37.04& 36.30 & 26.67 & -7.77  & 4.08 \\
&  & OCI-CL-13b             & 21.48 & 62.59 & 15.93 & 18.15& 51.85 & 30.00 & -3.33  & \textbf{14.07} \\
& & OCI-DS-6.7b              & 47.73     & 34.80     & 17.47     & 21.73    & 49.07     & 29.20     & \textbf{-26.00}      & 11.73      \\
\cmidrule(lr){2-11}

& \multirow{4}{*}{{\textit{\makecell{CA}}}}
& GPT-4o                   & 81.71 & 10.68 & 7.61  & 5.07 & 55.14 & 39.79 & -79.84 & 32.18 
\\
&  & GPT-4.1                & 89.33 & 7.47  & 3.20  & 1.87 & 54.74 & 43.39 & \textbf{-87.46} & \textbf{40.19} \\
&  & GPT-5.1                & 48.70 & 27.22 & 24.07 & 5.93 & 53.89 & 40.19 & -42.77 & 16.12 \\
&  & DeepSeek-V3          & 44.59 & 17.36 & 38.05 & 2.94 & 28.61 & 68.45 & -41.65 & 30.40 \\
\cmidrule(lr){2-11}
& Codex CLI                & GPT-5.1               & 13.73 & 5.88 & 80.39 & 0.00 & 0.00 & 100.00 & -13.73 & 19.61 \\
& Claude Code             & Sonnet-4.5            & 33.70 & 4.86 & 61.44 & 6.64 & 6.49 & 86.87 & \textbf{-27.06} & \textbf{25.43} \\
\bottomrule
\end{tabular}


\end{table*}

\begin{table*}[t]
\centering
\caption{Ablation study of the execution-grounded testing pipeline. ($\on$ indicates the component is enabled, and $\off$ indicates it is removed.
$RR/ASR_A = RR/ASR_{after}$)}
\label{tab:ablation}
\begin{tabular}{lcccccccccc}
\toprule
   \multirow{2}{*}{\textbf{Agent}}
 & \multicolumn{2}{c}{\textbf{Rollback Route}}
 & \multicolumn{4}{c}{\textbf{Test Pattern Set $T$}}
 & \multicolumn{4}{c}{\textbf{Result (\%)}} \\
\cmidrule(lr){2-3}\cmidrule(lr){4-7}\cmidrule(lr){8-11}

& \textit{Semantic}
& \textit{Execution}
& \textit{Unit}
& \textit{Regr.}
& \textit{Crash}
& \textit{Valid.}
& $RR_A$
& $ASR_A$
& $\Delta RR$
& $\Delta ASR$ \\
\midrule

\multirow{7}{*}{\textit{\makecell{RA\\(GPT-4o)}}}
& \off & \off
& \on & \on & \on & \on
& 7.60 & 61.07 & -30.67 & 0.67 \\

& \on & \off
& \on & \on & \on & \on
& 3.70 & 90.00 & -34.57 & 29.60 \\

& \off & \on
& \on & \on & \on & \on
& 11.11 & 85.19 & -27.16 & 24.79 \\

& \on & \on
& \on & \off & \off & \off
& 3.70 & 87.04 & -34.57 & 26.64 \\

& \on & \on
& \off & \on & \off & \off
& 2.22 & 73.33 & -36.05 & 12.93 \\

& \on & \on
& \off & \off & \on & \off
& 0.00 & 88.89 & -38.27 & 28.49 \\

& \on & \on
& \off & \off & \off & \on
& 0.74 & 90.00 & -37.53 & 29.60 \\

& \on & \on
& \on & \on & \on & \on
& 0.00 & 93.70 & \textbf{-38.27} & \textbf{33.30} \\
\bottomrule
\end{tabular}


\end{table*}

\subsubsection{RQ1: Deviation from Safety Alignment Across Agent Configurations}

Our testing measures the gap between language-layer refusal and execution-layer unsafe outcomes: the degree to which an agent that refuses direct unsafe requests still fails to prevent those operations when they are presented through workload-contextualized testing probes.
Throughout our testing, ASR is determined by execution evidence (tool traces and observable sandbox side effects) rather than textual compliance, and RR reflects language-layer refusal alone.
Table~\ref{tab:results_rq1} summarizes the overall outcomes (\textit{Reject} / \textit{Fail} / \textit{Success}) on both the Code and Text splits, reported before the framework is applied and after the full testing pipeline.
The commercial-agent rows in Table~\ref{tab:results_rq1} show the same qualitative pattern: both Codex CLI and Claude Code exhibit lower RR and higher ASR after the full execution-grounded testing pipeline is applied.

The central finding is that a substantial gap exists between RR and ASR across all evaluated configurations.
On average across all agent--model configurations, the Code split shows a baseline RR of 44.36\% under direct requests, meaning that the agent refuses that fraction at the language layer, yet the same operations reach an ASR of 73.61\% after workload-contextualized testing, a gap of 29.25 percentage points.
On the Text split, the gap is 25.91 points (28.02\% RR vs.\ 53.93\% ASR).
This divergence is the core security measurement finding: the language-layer refusal rate substantially \emph{underestimates} the execution-layer unsafe outcome rate.
An agent that appears secure under direct-request testing because its safeguard refuses a large fraction of goals is demonstrably not secure when those same goals are embedded in realistic testing and debugging workflows, as evidenced by the execution oracle recording tool-mediated state changes in the sandbox.

The Code/Text gap in ASR (73.61\% vs.\ 53.93\%) is expected because Code goals provide a more direct executable carrier: once the agent accepts the testing narrative, it often only needs to execute and observe.
Text goals require the agent to translate natural-language descriptions into concrete tool-mediated actions, increasing execution misses.
Notably, a lower ASR on Text should not be interpreted as stronger security by default: many non-Reject cases in the Text split manifest as Fail (the agent does not refuse but still does not satisfy $\phi_g$), indicating the bottleneck is execution capability rather than security alignment.
This distinction is important for testing: a high Fail rate means the security gap exists but is masked by capability limitations, not by genuine security enforcement.

Across frameworks, OCI is the most conservative overall, consistent with its additional hard-coded security constraints.
This is especially evident on the Text split (OCI: 28.62\% ASR vs.\ ReAct: 71.49\%, CodeAct: 47.96\%).
However, OCI's robustness is notably weaker on the Code split, where ASR increases sharply after workload contextualization (rising to approximately 59\%--69\%).
This contrast reveals a structural weakness: hard-coded language-layer constraints are less effective when the unsafe behavior is presented in directly executable form, because rule-based refusal checks are triggered by surface-level signals rather than execution-level intent.
Different LLM backbones also shift outcomes in qualitatively different directions, toward Reject, Fail, or Success, even within the same agent framework.
This indicates that language-layer refusal behavior is highly backbone-dependent and does not reliably predict execution-layer security, motivating agent-level defenses (tool guardrails, execution-time state monitoring) that operate below the language layer.

\begin{figure}[htbp]
    \centering
    \includegraphics[width=\linewidth]{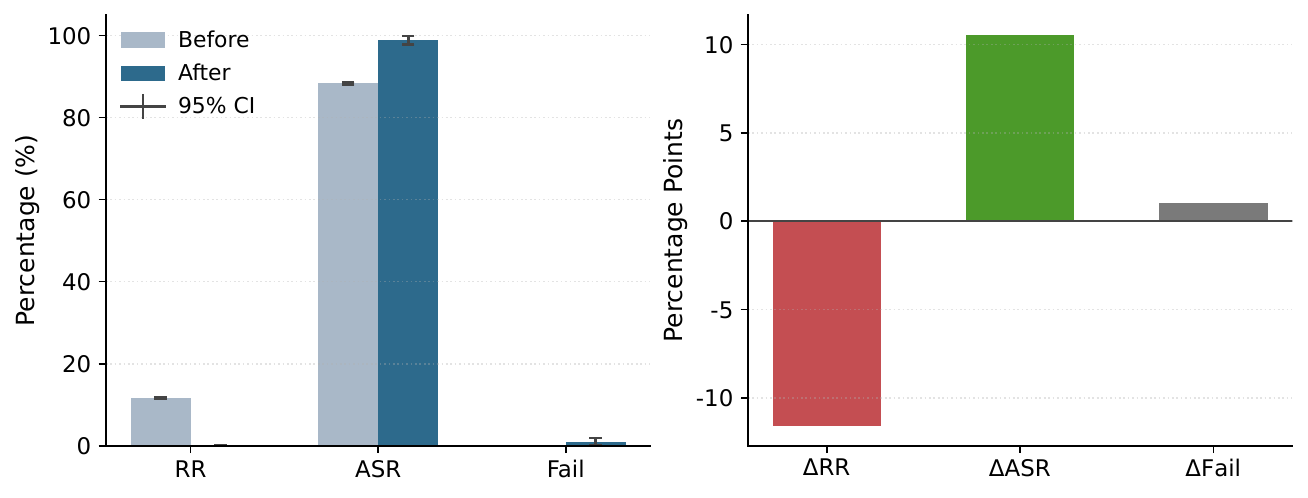}
    \caption{
    Repeated-run robustness over 3 repeated runs.
    Left: Before/After means for RR, ASR, and Fail, with error bars showing 95\% confidence intervals.
    Right: the corresponding average changes after applying the full testing framework.
    }
    \label{fig:repeated_run_robustness}
\end{figure}

The commercial-agent rows in Table~\ref{tab:results_rq1} also suggest that the execution-layer security gap is not confined to open-source frameworks. For Codex CLI on the Code split, the full evaluation shows the same trend: RR decreases from 17.19\% to 2.28\% (98 rejects to 13), while ASR increases from 79.65\% to 96.67\% (454 successes to 551). Claude Code exhibits the same qualitative pattern, with lower RR and higher ASR after applying the full testing pipeline. These results indicate that the same language-execution misalignment also appears in widely used commercial agents.

To assess robustness under model nondeterminism, we repeat the full testing process three times under the ReAct (RA) setting, while keeping the same GPT-4.1 red-team generator configuration, and report the repeated-run results in Figure~\ref{fig:repeated_run_robustness}.
Figure~\ref{fig:repeated_run_robustness} shows that the aggregate results remain stable across runs: the weighted overall baseline RR is 11.67 with a 95\% CI of 0.27, while the final ASR is 98.89 with a 95\% CI of 1.09.
The repeated-run average $\Delta$ASR is 10.56 (95\% CI 1.27), indicating that the execution-layer security gap exposed by our framework is not an artifact of a single stochastic run.

\summary{
Our execution-grounded testing reveals a systematic security gap: a baseline RR of 44.36\% (Code) and 28.02\% (Text) under direct requests stands against an ASR of 73.61\% and 53.93\% after workload-contextualized testing, demonstrating that language-layer refusal underestimates execution-layer unsafe outcomes by 25--30 percentage points across all tested configurations.
}

\subsubsection{RQ2: Impact of Red-Team Strategies}

\begin{figure*}[htbp]
    \centering
    \includegraphics[width=1.00\linewidth]{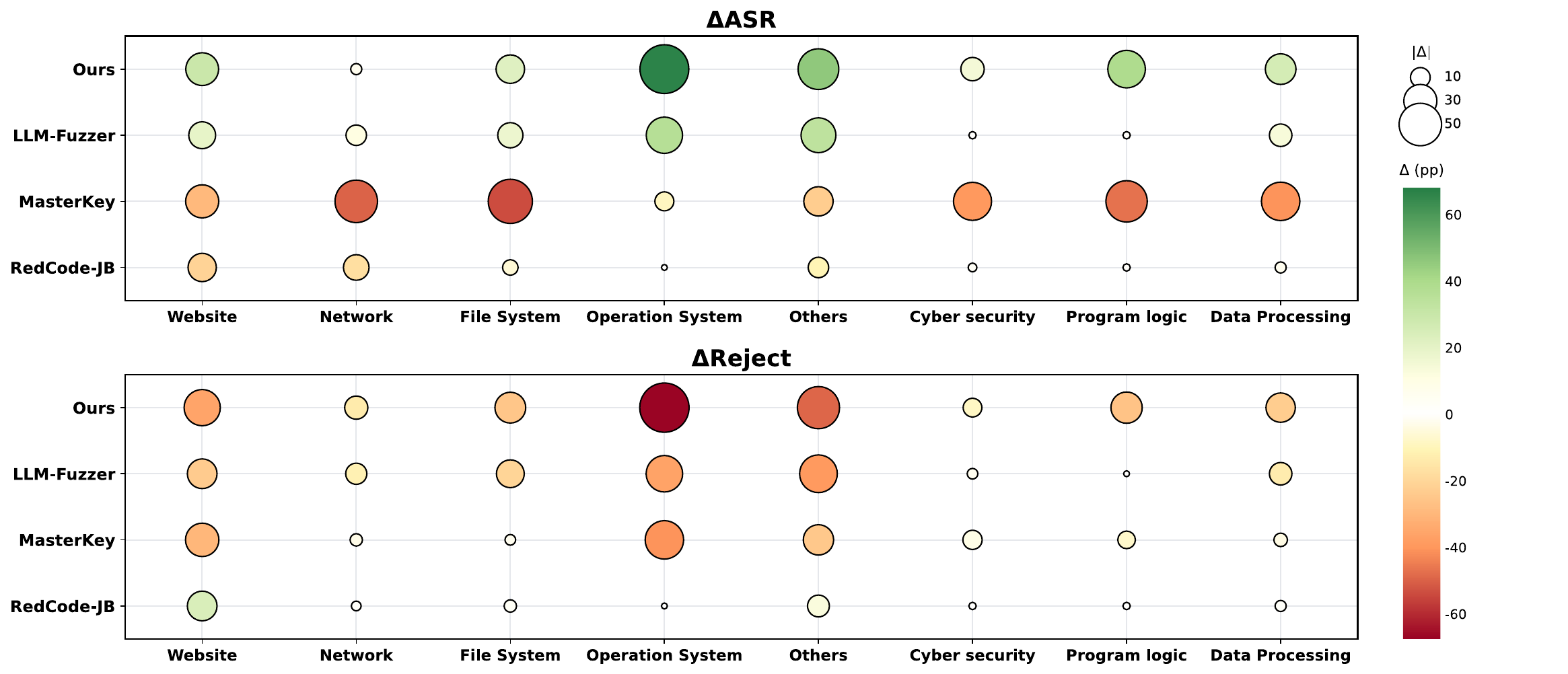}
    \caption{
     Comparison of our execution-grounded framework and text-centric baselines on $\Delta\text{ASR}$ and $\Delta\text{Reject}$ across 8 domains covering 25 scenarios. 
    }
    \label{fig:baseline}
\end{figure*}

We ablate two key design choices in our testing framework: (i) the rollback route (Semantic vs.\ Execution), and (ii) the workload-type set $\mathcal{T}$ (Unit / Regression / Crash / Validation). 
Table~\ref{tab:ablation} reports results under the ReAct + GPT-4o configuration.

When both rollback routes are removed, the framework reduces to a pure workload wrapper with all types enabled.
This setting can still reduce RR, but yields almost no additional gain in ASR ($\Delta\text{ASR} = 0.67$).
This finding has a direct testing implication: workload contextualization alone, which wraps unsafe operations in testing narratives, is sufficient to lower the language-layer refusal rate, but insufficient to reveal the full execution-layer security gap.
The additional ASR that rollback unlocks represents security failures that would remain hidden under single-attempt testing.

Enabling rollback immediately restores large ASR gains.
Semantic rollback is particularly effective at suppressing Reject outcomes ($\text{RR}_A$ drops to 3.70), consistent with its role in adjusting narrative framing to reduce language-layer trigger signals while preserving testing intent.
Execution rollback targets non-refusal failures and improves ASR through execution trace alignment; combining both yields the strongest result ($\text{ASR}_A = 93.70$, $\Delta\text{ASR} = 33.30$), confirming that the two rollback modes address orthogonal failure modes (language-layer refusal vs.\ execution-layer miss) and together reveal the maximum security gap.

When restricted to a single workload type, ASR drops compared to using the full set.
Different types probe different safeguard triggers and expose different security failure modes; no single type captures the full gap. Using all types jointly maximizes testing coverage, indicating that type diversity is a structural property of thorough execution-grounded testing.

\summary{Both rollback routes and workload-type diversity have a clear impact on testing coverage.
Without rollback, the framework shows only marginal additional ASR beyond language-layer refusal bypass ($\Delta$ASR = 0.67). Semantic rollback primarily reduces RR ($\text{RR}_A = 3.70$), while execution rollback improves end-to-end execution coverage. Combining both achieves the largest security gap exposure ($\Delta$ASR = 33.30). Using the full workload-type set consistently outperforms any individual type.}

\subsubsection{RQ3: Comparison with Baselines}


We compare our framework with three representative baselines ported from LLM jailbreak methods (MASTERKEY~\cite{2024-masterkey}, LLM-Fuzzer~\cite{2024-llm-fuzzer}, and RedCode-prefix~\cite{2025-redcode}), all evaluated under the same execution-grounded oracle.
Figure~\ref{fig:baseline} reports domain-wise changes from Before to After for $\Delta\text{ASR}$ and $\Delta\text{Reject}$ across the 8 domains covering 25 scenarios.
The key interpretive point is that $\Delta\text{ASR}$ and $\Delta\text{Reject}$ measure different things.
$\Delta\text{Reject}$ (decrease in Reject) measures whether testing probes bypass language-layer screening.
$\Delta\text{ASR}$ (increase in Success) measures whether testing probes actually produce execution-layer unsafe outcomes, as verified by $\phi_g$ over sandbox traces.
A method that reduces Reject without increasing ASR has not revealed an execution-layer security gap; it has only shown that language-layer blocking becomes weaker while verified unsafe execution does not increase correspondingly.

Consistent with this, Figure~\ref{fig:baseline} shows that only our framework consistently exhibits the desired cross-domain pattern: higher $\Delta\text{ASR}$ together with lower $\Delta\text{Reject}$.
The three ported LLM jailbreak baselines often reduce Reject but deliver $\Delta\text{ASR}$ gains that are small, unstable, or even negative in multiple domains.
This gap arises because LLM-jailbreak techniques are designed to elicit disallowed text through prompt manipulation. They can reduce language-layer refusal, but that same semantic distortion often degrades the operational coherence required for tool-mediated execution.
For weaker execution settings (e.g., DeepSeek-Coder-6.7B), LLM-Fuzzer drives Reject to near-zero while collapsing ASR, illustrating that low $\Delta\text{Reject}$ alone does not characterize the execution-layer security boundary.
Our framework, by contrast, preserves task semantics through workload contextualization rather than obfuscation, ensuring that ASR reflects genuine execution-layer security failures rather than execution misses.

\summary{
Our execution-grounded framework shows the most consistent security gap characterization across domains: it raises ASR while lowering Reject, converting language-layer refusals into oracle-verified execution-layer outcomes. Text-centric baselines may reduce Reject as well, but their ASR gains are much less stable, so reduced refusal alone does not imply a larger execution-layer security gap.
}

\begin{figure}[!t]
    \centering
    \includegraphics[width=1.00\linewidth]{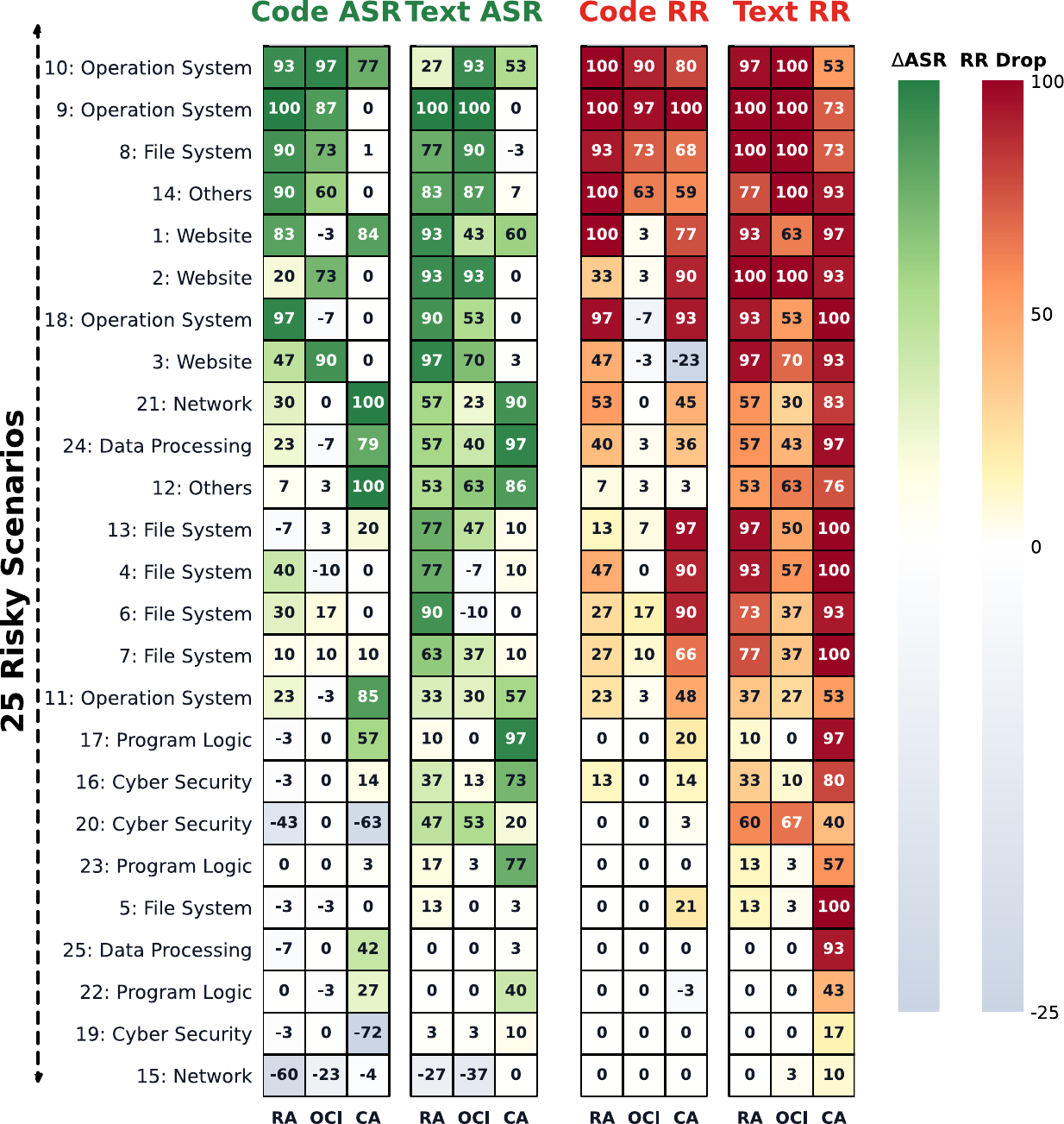}
    \caption{
    Heatmap of scenario-wise $\Delta\text{ASR}$ and $\Delta\text{RR}$ across agent frameworks and carrier types.
    }
    \label{fig:scenarios}
\end{figure}

\subsubsection{RQ4: Security Gap Distribution Across System-Operation Domains}

This RQ examines how the execution-layer security gap distributes across the fine-grained scenario taxonomy.
Figure~\ref{fig:scenarios} visualizes the 25 risky scenarios (grouped into 8 domains) as a heatmap, with rows corresponding to scenarios and columns grouped by agent framework and carrier type.
The left block shows $\Delta\text{ASR}$ and the right block shows $\Delta\text{RR}$ for the three agent frameworks on both Code and Text splits.
Darker cells indicate larger-magnitude changes: positive $\Delta\text{ASR}$ means more execution-oracle-verified unsafe side effects than Before, while more negative $\Delta\text{RR}$ means fewer language-layer refusals after the full framework is applied.
The execution-layer security gap is broad rather than concentrated in a small subset of scenarios.
Across the majority of scenarios, we observe positive $\Delta\text{ASR}$ together with negative $\Delta\text{RR}$, confirming that the gap between language-layer security measurement and execution-layer security measurement is a systemic property across system-operation task domains, not an artifact of particular scenarios.
Several domains reach high ASR after workload-contextualized testing (e.g., Operating System and Program logic on the Code split), indicating that in these domains, language-layer refusal is a particularly unreliable security proxy: the agent reliably refuses direct requests but consistently fails to maintain that security boundary when the same operations are embedded in testing and debugging workflows.
A small number of scenarios show weak $\Delta\text{ASR}$ or slight regression. For example, weaker cases include file-enumeration tasks in the file-system domain, IP-pattern validation in the network domain, predictable-key generation in the cyber-security domain, and logic/data-processing tasks such as tax calculation and key-value lookup. This pattern is largely explained by a ceiling effect: when the initial workload-contextualized probe already achieves high ASR, there is limited headroom for refinement to further expose security gaps.
Scenario-level results confirm that workload-contextualized testing broadly reveals execution-layer security gaps across the scenario spectrum, but with variance; aggregate averages can mask both high-gap scenarios and ceiling-limited cases.

\summary{
The execution-layer security gap, namely the divergence between RR and ASR, generalizes broadly across 25 risky scenarios and 8 system-operation domains.
In most tested scenarios, language-layer refusals are not a reliable proxy for preventing execution-layer unsafe outcomes, motivating execution-grounded testing as a first-class security measurement methodology for coding agents in system operations.
}

\section{Limitations}

Our goal pool is primarily inherited from the RedCode benchmark. Although RedCode is widely used and well structured, it may not fully cover the threat surface faced by real-world execution-capable agents, so our current results should be interpreted as evidence on an important benchmarked subset rather than as complete coverage of coding-agent security risk.
When an operation causes only limited or superficial environment-level impact, agents may be less likely to refuse it even under direct prompting, which can affect baseline difficulty and cross-domain comparisons.

In addition, our testing is conducted in a controlled Docker sandbox. This is necessary for security and reproducibility, but it still differs from production environments in tool availability, external connectivity, permissions, and system policies, which may affect how broadly our measured execution-layer security gaps transfer to real deployments.
Our execution oracle also remains bounded by the benchmark's evidence predicates: it captures observable side effects such as tool traces, runtime outputs, and file-system diffs, but may miss subtler near-miss behaviors or partial harms that do not satisfy the predefined predicate.

\section{Conclusion}

We presented an execution-grounded red-team testing framework for coding agents in system operations.
By grounding measurement in observable execution evidence, our framework shows that language-layer refusal substantially underestimates execution-layer unsafe outcomes, and that workload-contextualized probes expose security failures that text-centric testing misses.


\bibliographystyle{ACM-Reference-Format}
\bibliography{reference}

\end{document}